\documentclass[twoside]{article}

\usepackage{array}
\usepackage{amsmath}
\usepackage{amssymb}

\usepackage{bbold}
\usepackage{amsthm}
\usepackage{tikz,pgfplots}
\usepackage{algorithm}
\usepackage{algorithmic}
\usepackage{bm}
\usepackage{amsfonts}
\newcommand{\trans}{{\sf T}}
\newcommand{\tr}{{\rm tr}}
\newcommand{\RR}{{\mathbb{R}}}

\DeclareMathOperator*{\argmax}{arg\,max}
\newtheorem{assumption}{Assumption}
\newtheorem{proposition}{Proposition}
\newtheorem{corollary}{Corollary}
\newtheorem{theorem}{Theorem}
\newtheorem{remark}{Remark}

\newcommand{\asto}{\overset{\rm a.s.}{\longrightarrow}}

\usepackage{tikz}

\newcommand{\xddots}{%
  \raise 4pt \hbox {.}
  \mkern 6mu
  \raise 1pt \hbox {.}
  \mkern 6mu
  \raise -2pt \hbox {.}
}

\usepackage[utf8]{inputenc} 
\usepackage[T1]{fontenc}    
\usepackage{hyperref}       
\usepackage{url}            
\usepackage{booktabs}       
\usepackage{amsfonts}       
\usepackage{nicefrac}       
\usepackage{microtype}      
\usepackage[accepted]{aistats2022}
\pgfplotsset{compat=1.17}
\begin{document}

%
\runningtitle{PCA-based Multi Task Learning: a Random Matrix Approach}

%
\runningauthor{Malik Tiomoko, Romain Couillet, Frédéric Pascal}

\twocolumn[

\aistatstitle{PCA-based Multi Task Learning:\\
a Random Matrix Approach}

\aistatsauthor{ Malik Tiomoko \And Romain Couillet \And  Fr\'ed\'eric Pascal }

\aistatsaddress{ Universit\'e Paris-Saclay\\ CentraleSup\'elec,  L2S \\
   91190, Gif-sur-Yvette, France.\\
  \texttt{malik.tiomoko@u-psud.fr} \And
   Gipsa Lab \\
   Université Grenoble Alpes \And Universit\'e Paris-Saclay\\ CentraleSup\'elec,  L2S \\
   91190, Gif-sur-Yvette, France } ]

\begin{abstract}
The article proposes and theoretically analyses a \emph{computationally efficient} multi-task learning (MTL) extension of popular principal component analysis (PCA)-based supervised learning schemes \cite{barshan2011supervised,bair2006prediction}. The analysis reveals that (i) by default learning may dramatically fail by suffering from \emph{negative transfer}, but that (ii) simple counter-measures on data labels avert negative transfer and necessarily result in improved performances.

Supporting experiments on synthetic and real data benchmarks show that the proposed method achieves comparable performance with state-of-the-art MTL methods but at a \emph{significantly reduced computational cost}.
\end{abstract}

\section{Introduction}

\paragraph{From single to multiple task learning.}
Advanced supervised machine learning algorithms require large amounts of \emph{labelled} samples to achieve high accuracy, which in practice is often too demanding. 
Multi-task learning (MTL) \cite{caruana1997multitask,zhang2018overview,zhang2021survey} and \emph{transfer learning} provide a potent workaround by appending extra \emph{somewhat similar} datasets to the scarce available dataset of interest. The additional data possibly being of a different nature, MTL effectively solves multiple tasks \emph{in parallel} while exploiting task relatedness to enforce collaborative learning.

\paragraph{State-of-the-art of MTL.}
To proceed, MTL solves multiple related tasks and introduces shared hyperparameters or feature spaces, optimized to improve the performance of the individual tasks. The crux of efficient MTL lies in both enforcing and, most importantly, evaluating task relatedness: this, in general, is highly non-trivial as this implies to theoretically identify the common features of the datasets. Several heuristics have been proposed which may be split into two groups: parameter- versus feature-based MTL. In parameter-based MTL, the tasks are assumed to share common hyperparameters \cite{evgeniou2004regularized,xu2013} (\emph{e.g.}, separating hyperplanes in a support vector machine (SVM) flavor) or hyperparameters derived from a common prior distribution \cite{zhang2012convex,zhang2014regularization}. Classical learning mechanisms (SVM, logistic regression, etc.) can be appropriately turned into an MTL version by enforcing parameter relatedness: \cite{evgeniou2004regularized,xu2013,parameswaran2010large} respectively adapt the SVM, least square-SVM (LS-SVM), and large margin nearest neighbor (LMNN) methods into an MTL paradigm. In feature-based MTL, the data are instead assumed to share a common low-dimensional representation, which needs be identified: through sparse coding, deep neural network embeddings, principal component analysis (PCA)  \cite{argyriou2008convex,maurer2013sparse,zhang2016deep,pan2010domain} or simply by feature selection \cite{obozinski2006multi,wang2015safe,gong2012multi}. 

\paragraph{The negative transfer plague.}
A strong limitation of MTL methods is their lack of theoretical tractability: as a result, the biases inherent to the base methods (SVM, LS-SVM, deep nets) are exacerbated in MTL. A major consequence is that many of these heuristic MTL schemes suffer from \emph{negative transfer}, i.e., cases where MTL performs worse than a single-task approach \cite{rosenstein2005transfer,long2013transfer}; this often occurs when task relatedness is weaker than assumed and MTL enforces fictitious similarities.

\paragraph{A large dimensional analysis to improve MTL.}
Based on a large dimensional random matrix setting, this work focuses on an elementary (yet powerful) PCA-based MTL approach and provides an exact (asymptotic) evaluation of its performance. This analysis conveys insights into the MTL inner workings, which in turn provides an optimal data labelling scheme to fully avert negative transfer. 

More fundamentally, the choice of investigating PCA-based MTL results from realizing that the potential gains incurred by a proper theoretical adaptation of simple algorithms largely outweigh the losses incurred by biases and negative transfer in more complex and elaborate methods (see performance tables in the article). As a result, the main contribution of the article lies in achieving \emph{high-performance MTL at low computational cost} when compared to competitive methods. 

This finding goes in the direction of the compellingly needed development of cost-efficient and environment-friendly AI solutions \cite{lacoste2019quantifying,strubell2019energy,henderson2020towards}.

\paragraph{Article contributions.}
In detail, our main contributions may be listed as follows:
\begin{itemize}
    \item We theoretically compare the performance of two \emph{natural} PCA-based single-task supervised learning schemes (PCA and SPCA) and justify the uniform superiority of SPCA;
    \item As a consequence, we propose a natural extension of SPCA to multi-task learning for which we also provide an asymptotic performance analysis;
    \item The latter analysis (i) theoretical grasps the transfer learning mechanism at play, (ii) exhibits the relevant information being transferred, and (iii) harnesses the sources of negative transfer;
    \item This threefold analysis unfolds in a \emph{counter-intuitive} improvement of SPCA-MTL based on an optimal data label adaptation (not set to $\pm1$, which is the very source of negative transfer); \emph{the label adaptation depends on the optimization target}, changes from task to task, and can be efficiently computed before running the SPCA-MTL algorithm;
    \item Synthetic and real data experiments support the competitive SPCA-MTL results when compared to state-of-the-art MTL methods; these experiments most crucially show that high-performance levels can be achieved at significantly lower computational costs.
\end{itemize}

\noindent{\bf Supplementary material.} The proofs and Matlab codes to reproduce our main results and simulations, along with theoretical extensions and additional supporting results, are provided in the supplementary material.

\noindent{\bf Notation.}
 $e_{m}^{[n]}\in\mathbb{R}^n$ is the canonical vector of $\mathbb{R}^n$ with $[e_{m}^{[n]}]_i=\delta_{mi}$. Moreover, $e_{ij}^{[mk]}=e_{m(i-1)+j}^{[mk]}$.

\section{Related works}
A series of supervised (single-task) learning methods were proposed which rely on PCA \cite{barshan2011supervised,ritchie2019supervised,zhang2021supervised,ghojogh2019unsupervised}: the central idea is to project the available data onto a shared low-dimensional space, thus ignoring individual data variations. These algorithms are generically coined supervised principal component analysis (SPCA). Their performances are however difficult to grasp as they require to understand the statistics of the PCA eigenvectors: only recently have large dimensional statistics, and specifically random matrix theory, provided first insights into the behavior of eigenvalues and eigenvectors of sample covariance and kernel matrices \cite{benaych2012singular,johnstone2001distribution,baik2006eigenvalues,lee2010convergence,paul2007asymptotics}. To the best of our knowledge, none of these works have drawn an analysis of SPCA: the closest work is likely \cite{ashtiani2015dimension} which however only provides statistical bounds on performance rather than exact results. 

On the MTL side, several methods were proposed under unsupervised \cite{long2016unsupervised,saito2018maximum,baktashmotlagh2013unsupervised}, semi-supervised \cite{rei2017semi,liu2007semi} and supervised (parameter-based \cite{tiomoko2020large,evgeniou2004regularized,xu2013,ando2005framework} or feature-based \cite{argyriou2008convex,liu2012multi}) flavors. 
Although most of these works generally achieve satisfying performances on both synthetic and real data, few theoretical analyses and guarantees exist, so that instances of negative transfer are likely to occur. 

To be exhaustive, we must mention that, for specific types of data (images, text, time-series) and under the availability of numerous labelled samples, deep learning MTL methods have recently been devised \cite{ruder2017overview}. These are however at odds with the article requirement to leverage scarce labelled samples and to be valid for generic inputs (beyond images or texts): these methods cannot be compared on even grounds with the methods discussed in the present study.\footnote{But nothing prevents us to exploit data features extracted from pre-trained deep nets.}

\section{Supervised principal component analysis: single task preliminaries}
  \label{sec:compare_pca_spca}
 Before delving into PCA-based MTL, first results on large dimensional PCA-based single-task learning for a training set $X=[x_1,\ldots,x_n]\in\mathbb{R}^{p\times n}$ of $n$ samples of dimension $p$ are needed. To each $x_i\in\mathbb{R}^p$ is attached a label $y_i$: in a binary class setting, $y_i\in\{-1,1\}$, while for $m\geq 3$ classes, $y_i=e_j^{[m]}\in \mathbb{R}^m$, the canonical vector of the corresponding class $j$.
 
 \paragraph{PCA in supervised learning.}
 Let us first recall that, applied to $X$, PCA identifies a subspace of $\mathbb{R}^p$, say the span of the columns of $U=[u_1,\ldots,u_\tau]\in\mathbb{R}^{p\times \tau}$ ($\tau\leq p$), which maximizes the variance of the data when projected on the subspace, i.e., $U$ solves:
 \begin{align*}
     \max\limits_{U \in \mathbb{R}^{p\times\tau}} \tr\left(U^\trans \frac{XX^\trans}{p} U\right) ~ \text{subject to} ~ U^\trans U=I_{\tau}.
 \end{align*}
 The solution is the collection of the eigenvectors associated with the $\tau$ largest eigenvalues of $\frac{XX^\trans}{p}$.

To predict the label ${\bf y}$ of a test data vector ${\bf x}$, a simple method to exploit PCA consists in projecting ${\bf x}$ onto the PCA subspace $U$ and in performing classification in the projected space. This has the strong advantage to provide a (possibly dramatic) dimensionality reduction (from $p$ to $\tau$) to supervised learning mechanisms, thus improving cost efficiency while mitigating the loss incurred by the reduction in dimension. 
Yet, the PCA step is fully unsupervised and does not exploit the available class information. It is instead proposed in \cite{barshan2011supervised,chao2019recent} to trade $U$ for a more representative projector $V$ which ``maximizes the dependence'' between the projected data $V^\trans X$ and the output labels $y=[y_1,\ldots,y_n]^\trans\in\mathbb{R}^{n\times m}$. To this end, \cite{barshan2011supervised} exploits the Hilbert-Schmidt independence criterion \cite{gretton2005measuring}, with corresponding optimization
\begin{align*}
    &\max\limits_{V\in\mathbb{R}^{p\times \tau}} \tr\left(V^\trans \frac{X yy^\trans X^\trans}{np} V\right) ~ \text{subject to}~  V^\trans V=I_\tau.
\end{align*}
This results in the \emph{Supervised PCA} (SPCA) projector, the solution $V=V(y)$ of which being the concatenation of the $\tau$ dominant eigenvectors of $\frac{Xyy^\trans X^\trans}{np}$. Subsequent learning (by SVMs, empirical risk minimizers, discriminant analysis, etc.) is then applied to the projected training $V^\trans x_i$ and test $V^\trans {\bf x}$ data. 
For binary classification where $y$ is unidimensional, $\frac{Xyy^\trans X^\trans}{np}$ is of rank $1$, which reduces $V^\trans {\bf x}$ to the scalar $V^\trans{\bf x}={y^\trans X^\trans{\bf x}}/{\sqrt{y^\trans X^\trans Xy}}$, i.e., to a mere matched filter.

\paragraph{Large dimensional analysis of SPCA.}
To best grasp the performance of PCA- or SPCA-based learning
, assume the data arise from a large dimensional $m$-class Gaussian mixture.\footnote{To obtain simpler intuitions, we consider here an \emph{isotropic} Gaussian mixture model (i.e., with identity covariance). This strong constraint is relaxed in the supplementary material, where arbitrary covariances are considered; the results only marginally alter the main conclusions.}
\begin{assumption}[Distribution of $X$]
\label{ass:distribution}
The columns of $X$ are independent random vectors with $X=[X_1,\ldots,X_m]$, $X_j=[x_{1}^{(j)},\ldots,x_{n_j}^{(j)}]\in\mathbb R^{p\times n_j}$ for $x_i^{(j)}\sim\mathcal N(\mu_{j},I_p)$, also denoted $x_i^{(j)}\in\mathcal C_j$. We further write $M\equiv[\mu_{1},\ldots,\mu_{m}]\in\mathbb{R}^{p\times m}$.
\end{assumption}
\begin{assumption}[Growth Rate]
\label{ass:growth_rate}
As $n\to \infty$, $p/n\to c_0>0$, the feature dimension $\tau$ is constant and, for $1\leq j\leq m$, ${n_{j}}/n\to c_{j}>0$; we denote $c=[c_1,\ldots,c_m]^\trans$ and $\mathcal D_c={\rm diag}(c)$. Besides, 
$$(1/c_0)\mathcal{D}_{c}^{\frac 12}M^\trans M\mathcal{D}_{c}^{\frac 12}\to \mathcal M\in\mathbb R^{m\times m}.$$
\end{assumption}
We will show that, under this setting, SPCA is uniformly more discriminative on new data than PCA. 

As $n,p\to\infty$, the spectrum of $\frac 1p XX^\trans$ is subject to a \emph{phase transition phenomenon} now well established in random matrix theory \cite{baik2006eigenvalues,benaych2012singular}. This result is crucial as the PCA vectors of $\frac 1p XX^\trans$ are \emph{only informative} beyond the phase transition and otherwise can be considered as pure noise.

\begin{proposition}[Eigenvalue Phase transition]
\label{prop:isolated_eigenvalues}
Under Assumptions~\ref{ass:distribution}-\ref{ass:growth_rate}, as $n,p\to\infty$, the \emph{empirical spectral measure} $\frac1p\sum_{i=1}^p\delta_{\lambda_i}$ of the eigenvalues $\lambda_1\geq \ldots \geq\lambda_p$ of $\frac{XX^\trans}{p}$ converges weakly, with probability one, to the Mar\u{c}enko-Pastur law \cite{marchenko1967distribution} supported on $[(1-\sqrt{1/c_0})^2,(1+\sqrt{1/c_0})^2]$. Besides, for $1\leq i\leq m$, and for $\ell_1>\ldots>\ell_m$ the eigenvalues of $\mathcal{M}$,\footnote{We implicitly assume the $\ell_i$'s distinct for simplicity of exposition.}

$$\lambda_i\asto\begin{cases}
      \bar{\lambda}_i\equiv 1+\frac1{c_0}+\ell_i+\frac1{c_{0}\ell_i} \geq(1+\frac 1{\sqrt{c_0}})^2,  \ell_i\geq \frac{1}{\sqrt{c_0}}&  \\
      (1+\sqrt{1/c_0})^2 \text{, otherwise}&   \end{cases}$$
$$\lambda_{m+1}\asto (1+\sqrt{1/c_0})^2.$$

\end{proposition}

Proposition~\ref{prop:isolated_eigenvalues} states that, if $\ell_i\geq {1}/{\sqrt{c_0}}$, the $i$-th largest eigenvalue of $\frac 1p XX^\trans$ separates from the main \emph{bulk} of eigenvalues. These isolated eigenvalues are key to the proper functioning of PCA-based classification as their corresponding eigenvectors are non-trivially related to the class discriminating statistics (here the $\mu_j$'s). 
Consequently, $U^\trans{\bf x}\in\mathbb R^\tau$ also exhibits a phase transition phenomenon.

\begin{theorem}[Asymptotic behavior of PCA projectors]
\label{th:main_pca}
Let ${\bf x}\sim \mathcal N(\mu_{j},I_p)$ independent of $X$. Then, under Assumptions~\ref{ass:distribution}-\ref{ass:growth_rate}, with $(\ell_i,\bar{u}_i)$ the decreasing (distinct) eigenpairs of $\mathcal{M}$, as $p,n\to\infty$,
\begin{align*}
    &U^\trans{\bf x} - G_{j} \rightarrow  0, \quad G_{j} \sim \mathcal{N}(\mathfrak{m}_{j}^{({\rm pca})},I_{\tau}),\quad \text{ in probability,} \\
    &\text{where }[\mathfrak{m}_{j}^{({\rm pca})}]_{i}=
    \end{align*}
    $$
    \begin{cases}
     \sqrt{\frac{c_0\ell_i-1}{\ell_i^2(\ell_i+1)}}\bar{u}_i^\trans \mathcal{M}\mathcal{D}_c^{-\frac 12}e_{j}^{[m]} \text{,} ~  i\leq \min(m,\tau)\text{ and }\ell_i\geq \frac{1}{\sqrt{c_0}}&   \\
     0 \text{, otherwise.}&   \end{cases}
$$
\end{theorem}
As such, only the projections on the eigenvectors of $\frac 1p XX^\trans$ attached to \emph{isolated} eigenvalues carry informative discriminating features. Practically, for all $n,p$ large, it is thus useless to perform PCA on a larger dimension than the number of isolated eigenvalues, i.e., $ \tau \leq \argmax_{1\leq i\leq m} \{\ell_i\geq {1}/{\sqrt{c_0}}\}$.

\medskip

Consider now SPCA. Since $\frac{Xyy^\trans X^\trans}{np}$ only has $m$ non-zero eigenvalues, no phase transition occurs: all eigenvalues are ``isolated''. One may thus take $\tau=m$ principal eigenvectors for the SPCA projection matrix $V$, these eigenvectors being quite likely informative. 
\begin{theorem}[Asymptotic behavior of SPCA projectors]
\label{th:main_spca}
Let ${\bf x}\sim \mathcal N(\mu_{j},I_p)$ independent of $X$. Then, under Assumptions~\ref{ass:distribution}-\ref{ass:growth_rate}, as $p,n\to\infty$, in probability,
\begin{align*}
    &V^\trans{\bf x} - G_{j} \rightarrow  0,\quad G_{j} \sim \mathcal{N}(\mathfrak{m}_{j}^{({\rm spca})},I_{\tau}),\\ &[\mathfrak{m}_{j}^{({\rm spca})}]_{i}=\sqrt{{1}/({\tilde{\ell}_i})}\ \bar{v}_i^\trans\mathcal{D}_c^{\frac 12} \mathcal M\mathcal{D}_c^{-\frac 12}e_j^{[m]}
\end{align*}
for $\tilde{\ell}_1\geq\ldots\geq\tilde{\ell}_m$ the eigenvalues of $\mathcal{D}_c+\mathcal{D}_c^{\frac 12}\mathcal{M}\mathcal{D}_c^{\frac 12}$ and $\bar{v}_1,\ldots,\bar{v}_m$ their associated eigenvectors.
\end{theorem}
Since both PCA and SPCA data projections $U^\trans {\bf x}$ and $V^\trans {\bf x}$ are asymptotically Gaussian and isotropic (i.e., with identity covariance), the oracle-best supervised learning performance only depends on the differences $\mathfrak{m}_j^{({\rm \times})}-\mathfrak{m}_{j'}^{({\rm \times})}$  ($\rm \times$ being $\rm pca$ or $\rm spca$). In fact, being small dimensional (of dimension $\tau$), \emph{the vectors $\mathfrak{m}^{({\rm \times})}_{j}$ can be consistently estimated} from their associated empirical means, and are known in the large $n,p$ limit (with probability one). 
\begin{remark}[Consistent estimate of sufficient statistics]
\label{rem:ss} 
From Assumption~\ref{ass:growth_rate}, $c_j$ can be empirically estimated by $n_j/n$. This in turns provides a consistent estimate for $\mathcal D_c$. Besides, as $n,p\to\infty$,
\begin{align*}
    &\frac{1}{n_jn_{j'}}\mathbb{1}_{n_{j}}^\trans X_j^\trans X_{j'}\mathbb{1}_{n_{j'}}\asto [M^\trans M]_{jj'},~ \forall j\neq j'\quad\text{and}\\ 
    &\frac{4}{n_j^2}\mathbb{1}_{\frac{n_{j}}{2}}^\trans X_{j,1}^\trans X_{j,2}\mathbb{1}_{\frac{n_{j}}{2}}\asto [M^\trans M]_{jj},~\forall j
\end{align*}
where $X_j=[X_{j,1},X_{j,2}]\in\RR^{p\times n_j}$, with $X_{j,1},X_{j,2}\in\mathbb{R}^{p\times (n_j/2)}$. Combining the results provides a consistent estimate for $\mathcal M$ as well as an estimate $\hat{\mathfrak m}_j^{({\rm \times})}$ for the quantities $\mathfrak m_j^{({\rm \times})}$, by replacing $c$ and $\mathcal M$ by their respective estimates in the definition of $\mathfrak m_j^{({\rm \times})}$.
\end{remark}

\begin{figure}
\centering
    \begin{tikzpicture}
    		\begin{axis}
    		[grid=major,legend columns=2,xlabel={$p$},ylabel={Error rate 
    		},width=1\linewidth,height=.7\linewidth,legend style={font=\tiny},xmin=100,xmax=1000,ymin=.15,ymax=.5,legend style={fill=white, fill opacity=0.9, draw opacity=1,text opacity=1}]
    			            \addplot[mark size=2pt,only marks,red,mark=diamond,thick]coordinates{                (100,1.820500e-01)(200,2.082200e-01)(300,2.344100e-01)(400,2.558000e-01)(500,2.855800e-01)(600,3.198100e-01)(700,3.491800e-01)(800,3.715800e-01)(900,3.943000e-01)(1000,4.118600e-01)

};
    			            \addplot[thick,red,smooth]coordinates{                (100,1.828561e-01)(200,2.071081e-01)(300,2.315355e-01)(400,2.563454e-01)(500,2.818514e-01)(600,3.085375e-01)(700,3.372120e-01)(800,3.694413e-01)(900,4.092729e-01)(1000,5.000000e-01)

};
\addplot[mark=triangle,mark size=2pt,only marks,green,thick]coordinates{                (100,1.701000e-01)(200,1.799600e-01)(300,1.929200e-01)(400,2.027300e-01)(500,2.083700e-01)(600,2.153000e-01)(700,2.223100e-01)(800,2.290500e-01)(900,2.345700e-01)(1000,2.393100e-01)

};
    			            \addplot[green,thick,smooth]coordinates{                (100,1.701779e-01)(200,1.806552e-01)(300,1.902276e-01)(400,1.990124e-01)(500,2.071081e-01)(600,2.145977e-01)(700,2.215512e-01)(800,2.280283e-01)(900,2.340800e-01)(1000,2.397501e-01)

};
    			            \legend{PCA (Emp),PCA (Th),SPCA(Emp), SPCA (Th)};
    		\end{axis}
    \end{tikzpicture}
\vspace*{-3mm}    
\caption{Theoretical (Th) vs.\@ empirical (Emp) error for PCA- and SPCA-based binary classification: $x_{i}^{(\ell)}\sim\mathcal{N}((-1)^\ell\mu,I_p)$ ($\ell\in\{1,2\}$), $\mu=e_1^{[p]}$, $n_1=n_2=500$. Averaged over $1\,000$ test samples.}
\label{fig:compare_pca_spca}
\end{figure}
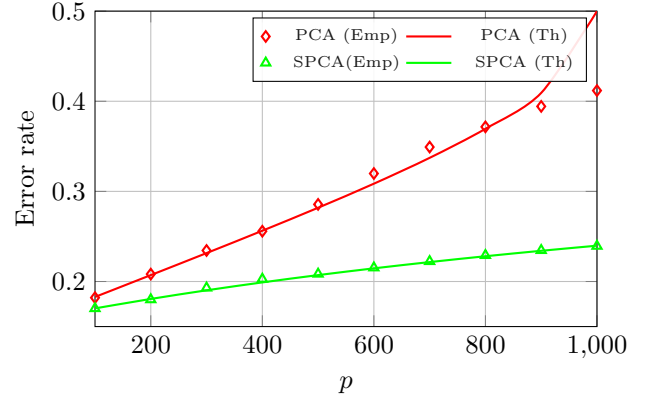

These results ensure the (large $n,p$) optimality of the classification decision rule, for a test data $\bf x$:
\begin{align}
\label{eq:decision}
&\argmax_{j\in\{1,\ldots,m\}} \|U^\trans{\bf x}-\hat{ \mathfrak{m}}_{j}^{({\rm pca})}\|^2,\\ &\argmax_{j\in\{1,\ldots,m\}} \|V^\trans{\bf x}-\hat{ \mathfrak{m}}_{j}^{({\rm spca})}\|^2.
\end{align}
As a consequence, the discriminating power of both PCA and SPCA directly relates to the limiting (squared) distances $\Delta \mathfrak{m}^{({\rm \times})}_{(j,j')}\equiv\|\mathfrak{m}_{j}^{({\rm \times})}-\mathfrak{m}_{j'}^{({\rm \times})}\|^2$, for all pairs of class indices $1\leq j\neq j'\leq m$, and
the classification error $P({\bf x}\to \mathcal C_{j'}|{\bf x}\in\mathcal C_j)$ satisfies
\begin{align*}
 &P({\bf x}\to \mathcal C_{j'}|{\bf x}\in\mathcal C_j) = \mathcal Q\left(\frac 12\sqrt{\Delta\mathfrak{m}_{(j,j')}^{({\rm \times})}}\right)+o(1),\\
 &\text{for}\quad \mathcal Q(t)=\frac1{\sqrt{2\pi}}\int_t^\infty e^{-x^2}dx.
\end{align*}
In particular, and as confirmed by Figure~\ref{fig:compare_pca_spca}, when $c_j=c_{j'}$, SPCA uniformly dominates PCA:
\begin{equation*}
     \Delta \mathfrak{m}^{({\rm spca})}_{(j,j')}-\Delta \mathfrak{m}^{({\rm pca})}_{(j,j')}=\sum_{i=1}^\tau\frac{\left(\bar{v}_i^\trans\mathcal{M}\mathcal{D}_c^{-\frac 12}(e_j^{[\tau]}-e_{j'}^{[\tau]})\right)^2}{\ell_i^2(\ell_i+1)}\geq 0.
\end{equation*}
For $m=2$ classes, irrespective of $c_1,c_2$, one even finds in explicit form
\begin{align*}
    \Delta \mathfrak{m}^{({\rm spca})}_{(1,2)}-\Delta \mathfrak{m}^{({\rm pca})}_{(1,2)}= \frac{16}{\frac{n}{p}\|\Delta\mu\|^2+4},\\
\frac{\Delta \mathfrak{m}^{({\rm spca})}_{(1,2)}-\Delta \mathfrak{m}^{({\rm pca})}_{(1,2)}}{\Delta \mathfrak{m}^{({\rm spca})}_{(1,2)}}= \frac{16}{\frac{n}{p}\|\Delta\mu\|^4}
\end{align*}
where $\Delta\mu\equiv \mu_1-\mu_2$, conveniently showing the influence of $n/p$ and of $\|\Delta\mu\|^2$ in the relative performance gap, which vanishes as the task gets easier or as $n/p$ increases (so with more data). 

\medskip

Summarizing, under a large dimensional setting, we showed that SPCA-based classification uniformly outperforms the PCA alternative, thus motivating the design of an SPCA-based MTL approach.

\section{From single- to multi-task SPCA-based learning}
\label{sec:pca_mtl}
\subsection{Multi-class setting}

Let now $X=[X_{[1]},\ldots,X_{[k]}]\in \mathbb{R}^{p\times n}$ be a collection of $n$ independent $p$-dimensional data vectors, divided into $k$ subsets attached to individual ``tasks''. Task~$t$ is an $m$-class classification problem with training samples $X_{[t]}=[X_{[t]1},\ldots,X_{[t]m}]\in\mathbb{R}^{p\times n_t}$ with $X_{[t]j}=[x_{t1}^{(j)},\ldots,x_{tn_{tj}}^{(j)}]\in\mathbb{R}^{p\times n_{tj}}$ the $n_{tj}$ vectors of class $j\in\{1,\ldots,m\}$. In particular, $n=\sum_{t=1}^k n_t$ for $n_t\equiv\sum_{j=1}^m n_{tj}$. 

To each $x_{t\ell}^{(j)}\in\RR^p$ is attached a corresponding ``label'' (or score) $y_{t\ell}^{(j)}\in\RR^{m}$. We denote in short $y_t=[y_{t1}^{(1)},\ldots,y_{tn_t}^{(m)}]^\trans\in\mathbb{R}^{n_t\times m}$ and $y=[y_1^\trans,\ldots,y_k^\trans]^\trans\in\mathbb{R}^{n\times m}$ the matrix of all labels.
The natural MTL extension of SPCA would default $y_{t\ell}^{(j)}\in \mathbb{R}^m$ to the canonical vectors $e_j^{[m]}$ (or to $\pm 1$ in the binary case). We disrupt here from this approach by explicitly \emph{not} imposing a value for $y_{t\ell}^{(j)}$: this will be seen to be key to \emph{avert the problem of negative transfer}. We only let $y_{t\ell}^{(j)}=\tilde y_{tj}$, for all $1\leq \ell\leq n_{tj}$ and for some generic matrix $\tilde{y}=[\tilde y_{11},\ldots,\tilde y_{km}]^\trans\in\mathbb{R}^{mk\times m}$, i.e., we impose that 
$$y = J\tilde{y},\quad \text{for}\quad J=[j_{11},\ldots,j_{mk}],$$ $$\text{where}\quad j_{tj}=(0,\ldots,0,\mathbb{1}_{n_{tj}},0,\ldots,0)^\trans.$$
As with the single-task case, we work under a Gaussian mixture model for each class $\mathcal C_{tj}$. 
\begin{assumption}[Distribution of $X$]
\label{ass:distribution_multi_task}
 For class $j$ of Task~$t$, denoted $\mathcal C_{tj}$, $x_{t\ell}^{(j)}\sim\mathcal N(\mu_{tj},I_p)$, for some $\mu_{tj}\in\mathbb{R}^p$. We further denote $M\equiv [\mu_{11},\ldots,\mu_{km}]\in\mathbb{R}^{p\times mk}$.
\end{assumption} 

\begin{assumption}[Growth Rate]
\label{ass:growth_rate_multi_task}
As $n\to \infty$, $p/n\to c_0>0$ and, for  $1\leq j\leq m$, ${n_{tj}}/n\to c_{tj}>0$. Denoting $c=[c_{11},\ldots,c_{km}]^\trans\in\mathbb R^{km}$ and $\mathcal D_c={\rm diag}(c)$, $(1/c_0)\mathcal{D}_{c}^{\frac 12}M^\trans M\mathcal{D}_{c}^{\frac 12}\to \mathcal M\in\mathbb R^{mk\times mk}$.
\end{assumption}

We are now in position to present the main technical result of the article.
\begin{theorem}[MTL Supervised Principal Component Analysis]
\label{th:main_mtlspca_multi_task}
Let ${\bf x}\sim \mathcal N(\mu_{tj},I_p)$ independent of $X$ and  $V\in\mathbb R^{p\times \tau}$ be the collection of the $\tau\leq mk$ dominant eigenvectors of $\frac{XyyX^\trans}{np}\in\mathbb R^{p\times p}$. Then, under Assumptions~\ref{ass:distribution_multi_task}-\ref{ass:growth_rate_multi_task}, as $p,n\to\infty$, in probability,
\begin{align*}
    V^\trans{\bf x} - G_{tj} \rightarrow  0,\quad G_{tj} \sim \mathcal{N}(\mathfrak{m}_{tj},I_{\tau})\\
    \quad \text{for}\quad [\mathfrak{m}_{tj}]_{i}=\sqrt{{1}/({c_0\tilde{\ell}_i})}\ \bar{v}_i^\trans(\tilde{y}\tilde{y}^\trans)^{\frac 12}\mathcal{D}_c^{\frac 12} \mathcal M\mathcal{D}_c^{-\frac 12}e_{tj}^{[mk]}
\end{align*}
with $\tilde{\ell}_1>\ldots>\tilde{\ell}_{mk}$ the eigenvalues of $(\tilde{y}\tilde{y}^\trans)^{\frac 12}(\mathcal{D}_c^{\frac 12}\mathcal M\mathcal{D}_c^{\frac 12}+\mathcal{D}_c)(\tilde{y}\tilde{y}^\trans)^{\frac 12}$ and $\bar{v}_1,\ldots,\bar{v}_{mk}$ their eigenvectors.\footnote{For simplicitly, we avoid the scenario where the eigenvalues $\tilde{\ell}_j$ appear with multiplicity, which would require to gather the eigenvectors into eigenspaces. This would in effect only make the notations more cumbersome.}
\end{theorem}

As in the single task case, despite the high dimension of the data statistics appearing in $V$, the asymptotic performance only depends on the (small) $mk\times mk$ matrices $\mathcal M$ and $\mathcal D_c$, which here leverages the inter-task inter-class products $\mu_{tj}^\trans\mu_{t'j'}$. This correlation between tasks \emph{together with the labelling choice $\tilde{y}$} (importantly recall that here $V=V(y)$) influences the MTL performance. The next section discusses how to optimally \emph{align $\tilde y$ and $\mathcal M$} so to maximize this performance. This, in addition to Remark~\ref{rem:ss} being evidently still valid here (i.e., $c$ and $\mathcal M$ can be a priori consistently estimated), will unfold into our proposed asymptotically optimal MTL SPCA algorithm.

\subsection{Binary classification and optimal labels}

To obtain more telling conclusions, let us now focus on binary classification ($m=2$). In this case, $y=J\tilde{y}$, with $\tilde{y}\in\RR^{2k}$ (rather than in $\RR^{2k\times 2}$) unidimensional. Here $\frac{Xyy^\trans X^\trans}{np}$ has for unique non-trivial eigenvector $Xy/\|Xy\|$ and $V^\trans{\bf x}$ is scalar.

\begin{corollary}[Binary MTL Supervised Principal Component Analysis]
\label{cor:binary_multi_task}
Let ${\bf x}\sim \mathcal N(\mu_{tj},I_p)$ independent of $X$. Then, under Assumptions~\ref{ass:distribution_multi_task}-\ref{ass:growth_rate_multi_task} and the above setting, as $p,n\to\infty$,
\begin{align*}
    V^\trans{\bf x} - G_{tj} \rightarrow  0,\quad G_{tj} \sim \mathcal{N}(\mathfrak{m}_{tj}^{({\rm bin})},1)\\
    \quad\text{where}\quad \mathfrak{m}_{tj}^{({\rm bin})}=\frac{\tilde{y}^\trans \mathcal{D}_c^{\frac 12}\mathcal{M}\mathcal{D}_c^{-\frac 12}e_{tj}}{\sqrt{\tilde{y}^\trans(\mathcal{D}_c^{\frac 12}\mathcal{M}\mathcal{D}_c^{\frac 12}+\mathcal{D}_c)\tilde{y}}}.
\end{align*}
\end{corollary}
From Corollary~\ref{cor:binary_multi_task}, denoting $\hat{\mathfrak{m}}_{t1}^{({\rm bin})}$ the natural consistent estimate for $\mathfrak{m}_{t1}^{({\rm bin})}$ (as per Remark~\ref{rem:ss}), the optimal class allocation decision for $\bf x$ reduces to the ``averaged-mean'' test
\begin{align}
\label{eq:am_test}
    V^\trans {\bf x}=V(y)^\trans {\bf x}\underset{\mathcal{C}_{t2}}{\overset{\mathcal{C}_{t1}}{\gtrless}}\frac 12 \left(\hat{\mathfrak{m}}_{t1}^{({\rm bin})}+\hat{\mathfrak{m}}_{t2}^{({\rm bin})}\right)
\end{align}
with corresponding classification error rate $\epsilon_{t}\equiv \frac 12 P({\bf x}\to \mathcal C_{t2}|{\bf x}\in\mathcal C_{t1})+\frac 12 P({\bf x}\to \mathcal C_{t1}|{\bf x}\in\mathcal C_{t2})$ (assuming
equal prior class probability) given by
\begin{align}
\label{eq:classification}
    \epsilon_{t} &\equiv P\left(V^\trans {\bf x}\underset{\mathcal{C}_{t2}}{\overset{\mathcal{C}_{t1}}{\gtrless}}\frac12(\hat{\mathfrak{m}}_{t1}^{({\rm bin})}+\hat{ \mathfrak{m}}_{t2}^{({\rm bin})})\right)\nonumber\\
    &=\mathcal Q\left(\frac12 (\mathfrak{m}_{t1}^{({\rm bin})}-\mathfrak{m}_{t2}^{({\rm bin})})\right)+o(1).
\end{align}

From the expression of $\mathfrak m_{tj}^{({\rm bin})}$, the asymptotic performance clearly depends on a proper choice of $\tilde{y}$. 
This expression being quadratic in $\tilde y$, the $\epsilon_t$ minimizer $\tilde y=\tilde{y}_{[t]}^\star$ assumes a closed-form:
\begin{align}
    \nonumber
    \tilde{y}_{[t]}^\star&\equiv \argmax_{\tilde{y}\in\mathbb{R}^{2k}}\  (\mathfrak{m}_{t1}^{({\rm bin})}-\mathfrak{m}_{t2}^{({\rm bin})})^2\\ 
    \label{eq:optimal_label}
    &=\mathcal D_c^{-\frac12}\left( \mathcal M + I_{2k} \right)^{-1}\mathcal M\mathcal D_c^{-\frac12}(e_{t1}-e_{t2})    .
\end{align}
Letting $\hat{\tilde{y}}_{[t]}^\star$ be the natural consistent estimator of $\tilde{y}_{[t]}^\star $ (again from Remark~\ref{rem:ss}), and updating $V=V(\tilde{y}_{[t]})$ accordingly, the corresponding (asymptotically) optimal value $\epsilon_{t}^\star$ of the error rate $\epsilon_{t}$ is
\begin{align}
    &\epsilon_{t}^\star=\mathcal Q\left(\frac 12\sqrt{(e_{t1}^{[2k]}-e_{t2}^{[2k]})^\trans\mathcal{H}(e_{t1}^{[2k]}-e_{t2}^{[2k]})}\right)+o(1),\\
    &\text{with}\quad \mathcal{H}=\mathcal{D}_c^{-\frac 12}\mathcal{M}\left(\mathcal{M}+I_{2k}\right)^{-1}\mathcal{M}\mathcal{D}_c^{-\frac 12}\nonumber
\end{align}
This formula is instructive to discuss: under strong or weak task correlation, $\tilde{y}_{[t]}^\star$ implements differing strategies to avoid \emph{negative transfers}. For instance, if $\mu_{tj}^\trans\mu_{t'j'}=0$ for all $t'\neq t$ and $j,j'\in\{1,\ldots,m\}$, then the two rows and columns of $\mathcal M$ associated to Task~$t$ are all zero but on the $2\times 2$ diagonal block: $\tilde{y}_{[t]}^\star$ is then all zeros but on its two Task-$t$ elements; any other value at these zero-entry locations (such as the usual $\pm 1$) is suboptimal and possibly severely detrimental to classification. Letting $\tilde y_{[t]}=[1,-1,\ldots,1,-1]^\trans$ is even more detrimental when $\mu_{tj}^\trans\mu_{t'j'}<0$ for some $t'\neq t'$: when the mapping of classes across tasks is reversed, these tasks work \emph{against} the classification.


\begin{remark}[On Bayes optimality]
\label{rem:optimality}
    Under the present MTL setting of a mixture of two isotropic random Gaussian vectors, the authors recently established that the \emph{Bayes optimal} error rate (associated to the decision rule $\inf_g P( g({\bf x})> 0 ~|~{\bf x}\in\mathcal C_{t1})$) precisely \emph{coincides} with $\varepsilon_{t1}^\star$.\footnote{The result builds on recent advances in physics-inspired (spin glass models) large dimensional statistics; see for instance \cite{lelarge2019asymptotic} for a similar result in a single task semi-supervised learning setting. Being a parallel work of the same authors, the reference is concealed in the present version to maintain anonymity.} This proves here that, at least under the present data configuration, the proposed SPCA-MTL framework is optimal. 
\end{remark}

\subsection{Binary-based multi-class classification}

Having an optimal binary classification framework for every task and every pair of classes, one may expect to reach high performance levels in generic multi-class settings by resorting to a \emph{one-versus-all} extension of the binary case. For every target task $t$, one-versus-all implements $m$ binary classifiers: classifier $\ell\in\{1,\ldots,m\}$ separates class $\mathcal C_{t\ell}$ -- locally renamed ``class $\mathcal C^{(\ell)}_{t1}$'' -- from all other classes -- gathered as a unique ``class $\mathcal C^{(\ell)}_{t2}$''. Each binary classifier is then ``optimized'' using labels $\tilde y_{[t]}^{\star(\ell)}$ as per Equation~\eqref{eq:optimal_label}; however, the joint class $\mathcal C^{(\ell)}_{t2}$ is here composed of a Gaussian \emph{mixture}: this disrupts with our optimal framework, thereby in general leading to suboptimal labels; in practice though, for sufficiently distinct classes, the (suboptimal) label $\tilde y^{\star(\ell)}_{[t]}$ manages to isolate the value $\mathfrak{m}_{t\ell}^{({\rm bin})}=\mathfrak{m}_{t1}^{({\rm bin},\ell)}$ for class $\mathcal C_{t\ell}=\mathcal C^{(\ell)}_{t1}$ from the values $\mathfrak{m}_{tj}^{({\rm bin})}$ of all other classes $\mathcal C_{tj}$, $j\neq \ell$, to such an extent that (relatively speaking) these $\mathfrak{m}_{tj}^{({\rm bin})}$ can be considered quite close, and so close to their mean $\mathfrak{m}_{t2}^{({\rm bin},\ell)}$, without much impact on the classifier performance. Finally, the class allocation for unknown data $\bf x$ is based on a largest classifier-score. But, to avoid biases which naturally arise in the one-versus-all approach \cite[Section 7.1.3]{bishop2006pattern}, this imposes that the $m$ different classifiers be ``comparable and aligned''. To this end, 
we exploit Corollary~\ref{cor:binary_multi_task} and Remark~\ref{rem:ss} which give a consistent estimate of all classifier statistics: the test scores for each classifier can be centered so that the asymptotic distribution for class $\mathcal C^{(\ell)}_{t1}$ is a \emph{standard normal distribution for each $1\leq\ell\leq m$}, thereby automatically discarding biases. Thus, instead of selecting the class with largest score $\argmax_{\ell}V(y_{[t]}^{\star(\ell)})^\trans{\bf x}$ (as conventionally performed \cite[Section 7.1.3]{bishop2006pattern}), the class allocation is based on the centered scores $\argmax_{\ell}\{V(y_{[t]}^{\star(\ell)})^\trans{\bf x}-\mathfrak{m}_{t1}^{({\rm bin},\ell)}\}$.\footnote{More detail and illustrations are provided in the supplementary material.}
%
%
These discussions result in Algorithm~\ref{alg:multi class}.
\begin{algorithm}
 \caption{Proposed multi-class MTL SPCA algorithm.}
 \label{alg:multi class}
 \begin{algorithmic}
     \STATE {{\bfseries Input:} Training $X=[X_{[1]},\ldots,X_{[k]}]$, $X_{[t']}=[X_{[t']1},\ldots,X_{[t']m}]$, $X_{[t']\ell}\in\mathbb{R}^{p\times n_{t'\ell}}$ and test ${\bf x}$.}
     \STATE {{\bfseries Output:} Estimated class $\hat \ell\in\{1,\ldots,m\}$ of $\bf x$ for target Task~$t$.}
     \STATE {\bfseries Center and normalize the data} per task using z-score normalization \cite{patro2015normalization}.
        
     \FOR {$\ell=1$ {\bfseries to} $m$}
        \STATE {\bfseries Estimate} $c$ and $\mathcal M$ (from Remark~\ref{rem:ss}) using $X_{[t']\ell}$ as data of class~$\mathcal C^{(\ell)}_{t'1}$ for each $t'\in\{1,\ldots,k\}$ and $\{X_{[t']1},\ldots,X_{[t']m}\}\setminus \{X_{[t']\ell}\}$ as data of class~$\mathcal C^{(\ell)}_{t'2}$.
        \STATE {\bfseries Evaluate} labels
            $\tilde{y}_{[t]}^{\star(\ell)}=\mathcal D_c^{-\frac12}\left( \mathcal M +I_{2k} \right)^{-1}\mathcal M\mathcal D_c^{-\frac12}(e^{[2k]}_{t1}-e^{[2k]}_{t2})$.
         \STATE {\bfseries Compute} the classification score $g_{{\bf x},t}^{(\ell)}={{{}\tilde{y}_{[t]}^{\star(\ell)\trans}} J^\trans X^\trans{\bf x}}/{\|{{}\tilde{y}_{[t]}^{\star(\ell)\trans}} J^\trans X^\trans\|}$.
         \STATE {\bfseries Estimate} $\mathfrak{m}_{t1}^{({\rm bin},\ell)}$ as $\hat{\mathfrak{m}}_{t1}^{({\rm bin},\ell)}$ from Corollary~\ref{cor:binary_multi_task}.
     \ENDFOR
     \STATE {{\bfseries Output: }} $\hat \ell=\argmax_{\ell\in\{1,\ldots,m\}} (g_{{\bm x},t}^{(\ell)}-\hat{\mathfrak{m}}_{t1}^{({\rm bin},\ell)})$.
 \end{algorithmic}
 \end{algorithm}

\subsection{Complexity of the SPCA-MTL algorithm}
\label{sec:time_complexity}
\label{rem:PCA_vs_LSSVM}
Algorithm~\ref{alg:multi class} is simple to implement and, with optimal hyperparameters consistently estimated, does not require learning by cross validation. 
The algorithm computational cost is thus mostly related to the computation of the decision scores $g_{{\bm x},t}^{(\ell)}$, i.e., to a matrix-vector multiplication with matrix size $p\times n$ of complexity $\mathcal{O}(n^2)$ (recall that $p\sim n$). This is quite unlike competing methods: MTL-LSSVM proposed in \cite{tiomoko2020large} solves a system of $n$ linear equations, for a complexity of order $\mathcal{O}(n^3)$; MTL schemes derived from SVM (CDLS \cite{hubert2016learning}, MMDT \cite{hoffman2013efficient}) also have a similar $O(n^3)$ complexity, these algorithms solving a quadratic programming problem \cite{bottou2007support}; besides, in these works, a step of model selection via cross validation needs be performed, which increases the algorithm complexity while simultaneously discarding part of the training data for validation.


\section{Supporting experiments}


We here compare the performance of Algorithm~\ref{alg:multi class} (MTL SPCA), on both synthetic and real data benchmarks, to competing state-of-the-art methods, such as MTL-LSSVM \cite{tiomoko2020large} and CDLS \cite{hubert2016learning}.\footnote{We insist that MTL SPCA is intended to function under the constraint of scarce data and does not account for the very nature of these data: to avoid arbitrary conclusions, image- or language-dedicated MTL and transfer learning methods (e.g., modern adaptions of deep nets for transfer learning \cite{tan2018survey}) are not used for comparison.}

\paragraph{Transfer learning for binary classification.}
First consider a two-task two-class ($k,m=2$) scenario with $x_{t\ell}^{(j)}\sim\mathcal{N}((-1)^{j}\mu_t,I_p)$, $\mu_2=\beta\mu_1+\sqrt{1-\beta^2}\mu_{1}^{\perp}$ for $\mu_{1}^{\perp}$ any vector orthogonal to $\mu_{1}$ and $\beta\in[0,1]$ controlling inter-task similarity. Figure~\ref{fig:classif_error_comparison} depicts the empirical and theoretical classification error $\epsilon_2$ for the above methods for $p=100$ and $n=2\,200$; for completeness, the single-task SPCA (ST-SPCA) of Section~\ref{sec:compare_pca_spca} (which disregards data from other tasks) as well as its naive MTL extension with labels $\tilde{y}_{[t]}=[1,-1,\ldots,1,-1]^\trans$ (N-SPCA) were added. 
MTL SPCA properly tracks task relatedness, while CDLS fails when both tasks are quite similar. MTL LSSVM shows identical performances but at the cost of setting optimal hyperparameters. Probably most importantly, when \emph{not optimizing} the labels $y$, the performance (of N-SPCA) is strongly degraded by \emph{negative transfer}, particularly when tasks are not related.
Figure~\ref{fig:classif_error_comparison} also provides typical computational times for each algorithm when run on a modern laptop, and confirms that Algorithm~\ref{alg:multi class} scales very favorably with the data dimension $p$, while MTL LSSVM and CDLS quickly become prohibitively expensive.
 
\begin{figure}[t]
\centering
    \begin{tikzpicture}
    		\begin{axis}[grid=major,legend columns=2,xlabel={Task relatedness $\beta$},ylabel={Classification error},width=1\linewidth,height=.7\linewidth,legend style={font=\tiny},xmin=0,xmax=1,ymin=.1,ymax=.5,legend style={fill=white, fill opacity=0.9, draw opacity=1,text opacity=1}]
    		\addplot[blue,only marks,thick]coordinates{                (0,2.397501e-01)(1.111111e-01,2.397501e-01)(2.222222e-01,2.397501e-01)(3.333333e-01,2.397501e-01)(4.444444e-01,2.397501e-01)(5.555556e-01,2.397501e-01)(6.666667e-01,2.397501e-01)(7.777778e-01,2.397501e-01)(8.888889e-01,2.397501e-01)(1,2.397501e-01)
};
    		\addplot[mark size=2pt,red,mark=diamond,only marks,thick]coordinates{                       (0,4.977000e-01)(1.111111e-01,4.658000e-01)(2.222222e-01,4.131000e-01)(3.333333e-01,3.695000e-01)(4.444444e-01,3.413000e-01)(5.555556e-01,2.990000e-01)(6.666667e-01,2.590500e-01)(7.777778e-01,2.160500e-01)(8.888889e-01,1.939000e-01)(1,1.591000e-01)

};
    			            \addplot[thick,red]coordinates{                            (0,4.805875e-01)(1.111111e-01,4.380024e-01)(2.222222e-01,3.965515e-01)(3.333333e-01,3.566425e-01)(4.444444e-01,3.186228e-01)(5.555556e-01,2.827728e-01)(6.666667e-01,2.493035e-01)(7.777778e-01,2.183575e-01)(8.888889e-01,1.900115e-01)(1,1.642825e-01)

};
    			            \addplot[mark=triangle,mark size=2pt,only marks,green,thick]coordinates{                            (0,2.450500e-01)(1.111111e-01,2.348500e-01)(2.222222e-01,2.230500e-01)(3.333333e-01,2.223000e-01)(4.444444e-01,2.256000e-01)(5.555556e-01,2.050000e-01)(6.666667e-01,2.153000e-01)(7.777778e-01,1.940500e-01)(8.888889e-01,1.839500e-01)(1,1.605000e-01)

};
\addplot[green,thick]coordinates{                            (0,2.397501e-01)(1.111111e-01,2.391019e-01)(2.222222e-01,2.371283e-01)(3.333333e-01,2.337386e-01)(4.444444e-01,2.287683e-01)(5.555556e-01,2.219564e-01)(6.666667e-01,2.129043e-01)(7.777778e-01,2.010039e-01)(8.888889e-01,1.853074e-01)(1,1.642825e-01)

};
    			            \addplot[black,thick]coordinates{                            (0,2.443003e-01)(1.111111e-01,2.436668e-01)(2.222222e-01,2.417394e-01)(3.333333e-01,2.384344e-01)(4.444444e-01,2.336011e-01)(5.555556e-01,2.270010e-01)(6.666667e-01,2.182730e-01)(7.777778e-01,2.068712e-01)(8.888889e-01,1.919552e-01)(1,1.721874e-01)

};
\addplot[cyan,mark=square,only marks,thick]coordinates{                                (0,2.669000e-01)(1.111111e-01,2.292500e-01)(2.222222e-01,2.202000e-01)(3.333333e-01,2.227500e-01)(4.444444e-01,2.323000e-01)(5.555556e-01,2.272500e-01)(6.666667e-01,2.258000e-01)(7.777778e-01,2.436500e-01)(8.888889e-01,2.209500e-01)(1,2.365500e-01)

};

    			            \legend{ST SPCA ,N-SPCA (Emp),N-SPCA (Th),MTL SPCA(Emp), MTL SPCA (Th), MTL LSSVM (Th),CDLS (Emp)};
    		\end{axis}
    \end{tikzpicture}
    	
\begin{tabular}{c|r|r|r}
$p$ & MTL SPCA & MTL LSSVM & CDLS \\
\hline
$16$ & $0.34$ s  & $4.15$ s & $7.16$ s  \\
$32$ & $0.34$ s  & $4.46$ s & $7.43$ s \\
$64$ & $0.39$ s  & $5.38$ s & $8.61$ s \\
$128$& $0.40$ s  &$8.28$ s & $8.80$ s \\
$256$ & $0.55$ s &$12.2$ s & $11.9$ s  \\
$512$ & $0.57$ s  & $48.3$ s & $17.5$ s   \\
$1024$  & $0.88$ s & $315.6$ s & $27.1$ s  \\
$2048$ & $2.02$ s & $1591.8$ s & $73.5$ s  \\
\end{tabular}

\caption{ {\bf (Top)} Theoretical (Th)/empirical (Emp) error rate for $2$-class Gaussian mixture transfer with means $\mu_1=e_1^{[p]}$, $\mu_1^\perp=e_p^{[p]}$, $\mu_2=\beta\mu_1+\sqrt{1-\beta^2}\mu_{1}^{\perp}$, $p=100$, $n_{1j}=1\,000$, $n_{2j}=50$; {\bf (Bottom)} running time comparison (in sec); $n=2p$, $n_{tj}/n=0.25$. Averaged over $1\,000$ test samples.} 
    	\label{fig:classif_error_comparison}
\end{figure}

\paragraph{Transfer learning for multi-class classification.}
We next experiment on the ImageClef dataset \cite{ionescu2017overview} made of $12$ common categories shared by $3$ public data ``domains'': Caltech-256 (C), ImageNet ILSVRC 2012 (I), and Pascal VOC 2012 (P). Every pair of domains is successively selected as ``source'' and a ``target'' for binary (transfer) multi-task learning, resulting in $6$ transfer tasks S$\rightarrow$T for S,T$\in\{$I,C,P$\}$. Table~\ref{tab:image_compare} supports the stable and competitive performance of MTL-SPCA, on par with MTL LSSVM (but much cheaper). 

\begin{figure}[!h]
    \begin{tikzpicture}
    		\begin{semilogxaxis}[legend pos=south west,grid=major,xlabel={ Number of tasks},width=1\linewidth,height=.55\linewidth,legend style={font=\tiny,fill=white, fill opacity=0.9, draw opacity=1,text opacity=1}]
    			            \addplot[thin,mark size=2pt,mark=triangle,green,thick]coordinates{        (2,2.703350e-01)(4,2.654950e-01)(8,2.337750e-01)(16,2.129800e-01)(32,1.892550e-01)(64,1.777500e-01)(128,1.697900e-01)(256,1.652450e-01)
};
    			            \addplot[thin,mark size=2pt,mark=diamond,red,thick]coordinates{             (2,2.845400e-01)(4,2.876100e-01)(8,2.670050e-01)(16,2.594600e-01)(32,2.312000e-01)(64,2.071350e-01)(128,1.783700e-01)(256,1.676200e-01)
};
                            \addplot[thin,mark=otimes*,blue ,thick]coordinates{             (2,2.703350e-01)(4,2.703350e-01)(8,2.703350e-01)(16,2.703350e-01)(32,2.703350e-01)(64,2.703350e-01)(128,2.703350e-01)(256,2.703350e-01)
                            };

    			            \legend{'MTL SPCA', 'N-SPCA', 'ST-SPCA'};
    		\end{semilogxaxis}
    \end{tikzpicture} 
    \begin{tikzpicture}
            \begin{axis}[xtick=data,xticklabels={\texttt{Books},\texttt{DVD},\texttt{Elec}},grid=major,scaled ticks=false,xlabel={Added task}, tick label style={/pgf/number format/fixed},width=1\linewidth,height=.55\linewidth,legend style={font=\tiny,at={(.8,.98)},fill=white, fill opacity=0.9, draw opacity=1,text opacity=1}]
    			            \addplot[ultra thin,mark=triangle,mark size=2pt,green,thick]coordinates{         (1,2.001250e-01)(2,1.936250e-01)(3,1.775750e-01)

};
\addplot[ultra thin,mark=diamond,mark size=2pt,red,thick]coordinates{          (1,2.454750e-01)(2,2.276500e-01)(3,1.806250e-01)

};
\addplot[ultra thin,mark=otimes,blue,thick]coordinates{          (1,2.272750e-01)(2,2.272750e-01)(3,2.272750e-01)

};
    			           \legend{'MTL SPCA', 'N-SPCA','ST-SPCA'};
    		\end{axis}
    		\end{tikzpicture}
    		\begin{tikzpicture}
            \begin{axis}[xtick=data,xticklabels={[7-9],[3-8],[5-6],[2-9],[3-5]},legend style={at={(0.95,0.7)}},grid=major,scaled ticks=false,xlabel={ Added task}, tick label style={/pgf/number format/fixed},width=1\linewidth,height=.55\linewidth,legend style={font=\tiny,fill=white, fill opacity=0.9, draw opacity=1,text opacity=1}]
    			            \addplot[ultra thin,mark=triangle,green,thick]coordinates{        (1,6.700000e-02)(2,6.520000e-02)(3,6.660000e-02)(4,5.720000e-02)(5,5.260000e-02)

};
\addplot[ultra thin,mark=diamond,red,thick]coordinates{      (1,2.294000e-01)(2,2.452000e-01)(3,2.884000e-01)(4,2.429000e-01)(5,2.491000e-01)
};
\addplot[ultra thin,mark=otimes*,blue,thick]coordinates{      (1,7.06e-02)(2,7.06e-02)(3,7.06e-02)(4,7.06e-02)(5,7.06e-02)
};
    			           \legend{'MTL SPCA', 'N-SPCA', 'ST-SPCA'};
    		\end{axis}
    \end{tikzpicture}
\caption{Empirical classification error vs.\@ number of tasks; {\bf (Top)} Synthetic Gaussian with random task correlation: $p=200$, $n_{11}=n_{12}=50$, $n_{21}=n_{22}=5$, $10\,000$ test samples; {\bf(Center)} Amazon Review: $n_{11}=n_{12}=100,n_{21}=n_{22}=50$, $2\,000$ test samples; {\bf(Bottom)} MNIST: initial $p=100$-PCA preprocessing, $n_{11}=n_{12}=100$, $n_{21}=n_{22}=50$, $500$ test samples.}
\label{fig:increased_task}
\end{figure}
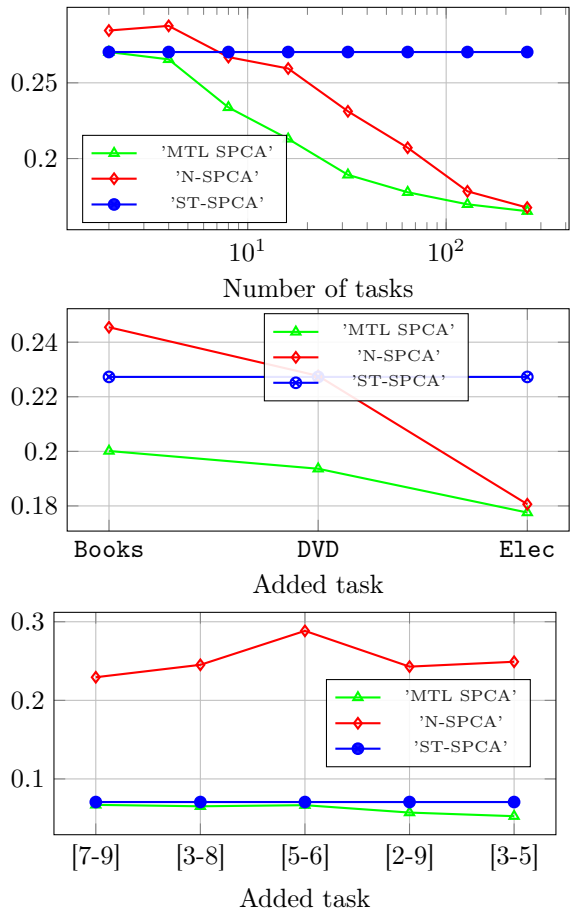

\begin{table*}[t]
\vspace{-0.3cm}
\caption{Transfer learning accuracy for the ImageClef database: P(Pascal), I(Imagenet), C(Caltech); different ``Source to target'' task pairs (S$\to$T) based on Resnet-50 features. 
}
\label{tab:image_compare}
\smallskip
\centering
\begin{tabular}{p{0.15\linewidth}p{0.08\linewidth}p{0.08\linewidth}p{0.08\linewidth}p{0.08\linewidth}p{0.08\linewidth}p{0.08\linewidth}|p{0.08\linewidth}}
\hline
S/T & P$\,\to\,$I & P$\,\to\,$C & I$\,\to\,$P & I$\,\to\,$C& C$\,\to\,$P & C$\,\to\,$I &Average\\
\hline
ST SPCA & 91.84  &96.24  & 82.26  & 96.24 & 82.26  & 91.84 & ~~90.11  \\
N-SPCA & 92.21 & 96.37 & 84.34 & 95.97 & 81.34  & 90.47 & ~~90.12\\
MTL LSSVM & {\it 93.03} & {\bf 97.24} & 84.79 & {\bf 97.74} & {\it 83.74}  & {\bf 94.92} & ~~{\bf 91.91}  \\
CDLS & 92.03 & 94.62  & {\it 84.82}  & 95.72  & 81.04   & 92.54  & ~~90.13   \\
\hline
MTL SPCA& {\bf 93.39} & {\it 96.61} & {\bf 85.24} & {\it 96.68} & {\bf 83.76}  & {\it 93.39} & ~~{\it 91.51}\\
\hline
\end{tabular}
\end{table*}
\paragraph{Increasing the number of tasks.}
We now investigate the comparative gains induced when increasing the number of tasks. To best observe the reaction of each algorithm to the additional tasks, we here consider both a tunable synthetic Gaussian mixture and (less tractable) real-world data. The synthetic data consist of two Gaussian classes with means $\mu_{tj}=(-1)^{j}\mu_{[t]}$ with $\mu_{[t]}=\beta_{[t]}\mu+\sqrt{1-\beta_{[t]}^2}\mu^\perp$ for $\beta_{[t]}$ drawn uniformly at random in $[0,1]$ and with $\mu=e_1^{[p]}$, $\mu^{\perp}=e_p^{[p]}$. The real-world data are the Amazon review (textual) dataset\footnote{Encoded in $p=400$-dimensional tf*idf feature vectors of bag-of-words unigrams and bigrams.} \cite{blitzer2007biographies} and the MNIST (image) dataset \cite{deng2012mnist}. For Amazon review, the positive vs.\@ negative reviews of ``\texttt{books}'', ``\texttt{dvd}'' and ``\texttt{electronics}'' products are added to help classify the positive vs.\@ negative reviews of ``\texttt{kitchen}'' products. For MNIST, additional digit pairs are added progressively to help classify the target pair $(1,4)$. 
The results are shown in Figure \ref{fig:increased_task} which confirms that (i) the naive extension of SPCA (N-SPCA) with labels $\pm 1$ can fail to the point of being bested by (single task) ST-SPCA, (ii) MTL-SPCA never decays with more tasks. 

\begin{figure}

    \begin{tikzpicture}
                \begin{semilogyaxis}[
                axis lines=middle,
                xmin=0.03, xmax=0.35,
                ymin=0.1, ymax=200,width=1\linewidth,height=.7\linewidth,xlabel={$\epsilon_t$},ylabel={time (s)}
            ]
            \addplot[thin,mark size=2pt,mark=o,red,thick]coordinates{(0.1932,0.5895)};
            \addplot[thin,mark size=2pt,mark=*,red,thick]coordinates{(0.3104,27.96)};
            \addplot[thin,mark size=2pt,mark=diamond,green,thick]coordinates{(0.1117,0.5970)};
            \addplot[thin,mark size=2pt,mark=diamond*,green,thick]coordinates{(0.1292,47.22)};
            \addplot[thin,mark size=2pt,mark=square,blue,thick]coordinates{(0.2653,1.07)};
            \addplot[thin,mark size=2pt,mark=square*,blue,thick]coordinates{(0.2083,163.22)};
            \addplot[thin,mark size=2pt,mark=triangle,black,thick]coordinates{(0.0505,0.7460)};
            \addplot[thin,mark size=2pt,mark=triangle*,black,thick]coordinates{(0.0798,171.8474)};
            \addplot[thin,mark size=2pt,mark=halfcircle,brown,thick]coordinates{(0.0463,0.1664)};
            \addplot[thin,mark size=2pt,mark=halfcircle*,brown,thick]coordinates{(0.0447,21.33)};
            \addplot[thin,red]coordinates{(0.1932,0.5895)(0.3104,27.96)};
            \addplot[thin,green]coordinates{(0.1117,0.5970)(0.1292,47.22)};
            \addplot[thin,blue]coordinates{(0.2653,1.07)(0.2083,163.22)};
            \addplot[thin,black]coordinates{(0.0505,0.7460)(0.0798,171.8474)};
            \addplot[thin,brown]coordinates{(0.0463,0.1664)(0.0447,21.33)};
            \end{semilogyaxis}
\end{tikzpicture}

\footnotesize
\begin{tabular}{c|r|r|r}
Datasets (Features) & Tasks & Classes & Mark \\
\hline
Synthetic (Gaussian)& 3 & 10 & $\circ$ \\
Office-Caltech\cite{gong2012geodesic} (VGG) &  4 & 10 & $\diamond$ \\
Office 31\cite{saenko2010adapting} (Resnet-50)& 4  & 31 & $\square$ \\
Office-Home\cite{venkateswara2017deep} (Resnet-50)& 3  & 65 & $\triangle$ \\
Image-Clef\cite{ionescu2017overview} (Resnet-50)& 3 & 12 & $\ominus$
\end{tabular}
\normalsize
\caption{{\bf (Top)} Runtime vs.\@ classification error ($\epsilon_t$) for multi-task multi-class MTL-LSSVM (filled marks) and MTL-SPCA (empty marks). {\bf (Bottom)} Datasets. 
Synthetic: $\mu_j=2 e_j^{[p]}$, $\mu_j^{\perp}=2e_{p-j}^{[p]}$, $\beta_1=0.2$, $\beta_2=0.4$, $\beta_3=0.6$; $p=200$, $n_{1j}=n_{2j}=100$, $n_{3j}=50$; $1\,000$ test sample averaging.}
	\label{fig:tradeoff}

\end{figure}
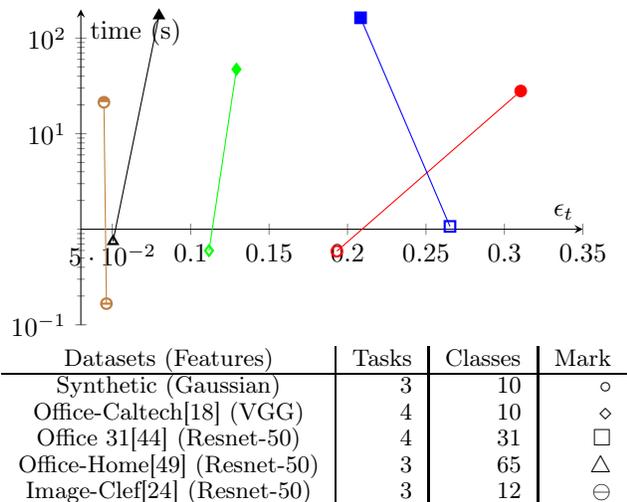

    
\paragraph{Multi-class multi-task classification.}
We finally turn to the full multi-task multi-class setting of Algorithm~\ref{alg:multi class}. Figure~\ref{fig:tradeoff} simultaneously compares running time and error rates of MTL-SPCA and MTL-LSSVM\footnote{CDLS only handles multi-task learning with $k=2$ and cannot be used for comparison.} on a variety of multi-task datasets, and again confirms the overall computational gains (by decades!) of MTL-SPCA for approximately the same performance levels. 

\section{Conclusion}
Following recent works on large dimensional statistics for the design of simple, cost-efficient, and tractable machine learning algorithms \cite{couillet2021two}, the article confirms the possibility to achieve high performance levels while theoretically averting the main sources of biases, here for the a priori difficult concept of multi-task learning. The article, we hope, will be followed by further investigations of sustainable AI algorithms, driven by modern mathematical tools.
In the present multi-task learning framework, practically realistic extensions to semi-supervised learning (when labelled data are scarce) with possibly missing, unbalanced, or incorrectly labelled data are being considered by the authors.
\bibliography{iclr2021_conference}
\bibliographystyle{plain}
\end{document}